\pdfoutput=1

\documentclass[11pt]{article}

\usepackage[preprint]{acl}

\usepackage{times}
\usepackage{latexsym}

\usepackage[T1]{fontenc}

\usepackage[utf8]{inputenc}

\usepackage{microtype}

\usepackage{inconsolata}

\usepackage{graphicx}
\usepackage{soul}
\usepackage{url}
\usepackage{hyperref}
\usepackage{caption}
\usepackage{amsmath}
\usepackage{array} 
\usepackage{amsthm}
\usepackage{booktabs}
\usepackage{algorithm}
\usepackage{algorithmic}
\usepackage{multirow}
\usepackage{makecell}

\usepackage{algorithmic}
\usepackage{color, colortbl}
\definecolor{Gray}{gray}{0.9}
\usepackage{tcolorbox}
\usepackage{soul}
\usepackage{listings}

\floatstyle{ruled}
\newfloat{listing}{tb}{lst}{}
\floatname{listing}{Listing}

\usepackage{lipsum}

\usepackage{xcolor}
\usepackage{xspace}
\usepackage{multirow}
\usepackage{multicol}
\usepackage{utfsym}

\usepackage[capitalize]{cleveref}

\def\benchname{ProactiveVideoQA\xspace}
\def\metricname{PAUC\xspace}

\title{\benchname: A Comprehensive Benchmark Evaluating Proactive Interactions in Video Large Language Models}

\author{
 \textbf{Yueqian Wang\textsuperscript{1}},
 \textbf{Xiaojun Meng\textsuperscript{2}},
 \textbf{Yifan Wang\textsuperscript{3}},
 \textbf{Huishuai Zhang\textsuperscript{1,4}},
 \textbf{Dongyan Zhao\textsuperscript{1,4}},
\\
\\
 \textsuperscript{1}Wangxuan Institute of Computer Technology, Peking University
\\
 \textsuperscript{2}Huawei Noah’s Ark Lab
\\
 \textsuperscript{3}School of Intelligence Science and Technology, University of Science and Technology Beijing
\\
 \textsuperscript{4}National Key Laboratory of General Artificial Intelligence
\\
 \small{
   \textbf{Correspondence:} \href{mailto:zhanghuishuai@pku.edu.cn}{zhanghuishuai@pku.edu.cn}, \href{mailto:zhaodongyan@pku.edu.cn}{zhaodongyan@pku.edu.cn}
 }
}

\usepackage{xcolor}

\begin{document}
\maketitle
\begin{abstract}
With the growing research focus on multimodal dialogue systems, the capability for proactive interaction is gradually gaining recognition. As an alternative to conventional turn-by-turn dialogue, users increasingly expect multimodal systems to be more initiative, for example, by autonomously determining the timing of multi-turn responses in real time during video playback.
To facilitate progress in this emerging area, we introduce \benchname, the first comprehensive benchmark to evaluate a system's ability to engage in proactive interaction. 
Since model responses are generated at varying timestamps, we further propose \metricname, the first metric that accounts for the temporal dynamics of model responses. This enables a more accurate evaluation of systems operating in proactive settings.
Through extensive benchmarking of various baseline systems on \benchname and a user study of human preferences, we show that \metricname is in better agreement with human preferences than traditional evaluation metrics, which typically only consider the textual content of responses. These findings demonstrate that \metricname provides a more faithful assessment of user experience in proactive interaction scenarios. \footnote{Project homepage: \href{https://github.com/yellow-binary-tree/ProactiveVideoQA}{https://github.com/yellow-binary-tree/ProactiveVideoQA}}
\end{abstract}

\section{Introduction} \label{sec:intro}

\begin{figure*}
    \centering
    \begin{minipage}{0.7\textwidth}
        \centering
        \includegraphics[width=\linewidth]{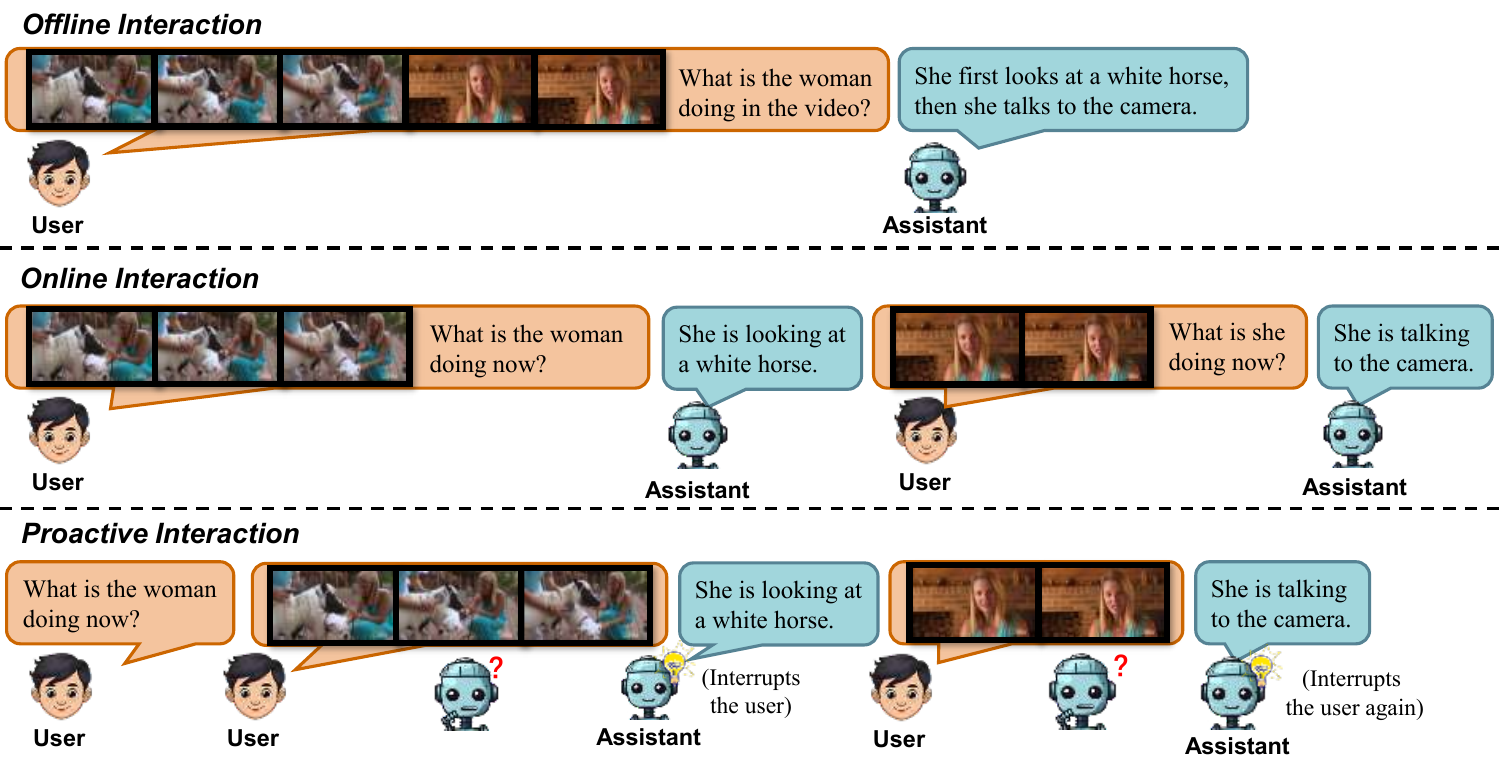}
        \caption{A demonstration of offline, online and proactive interaction.}
        \label{fig:interaction}
    \end{minipage}
    \hfill
    \begin{minipage}{0.28\textwidth}
        \centering
        \includegraphics[width=\linewidth]{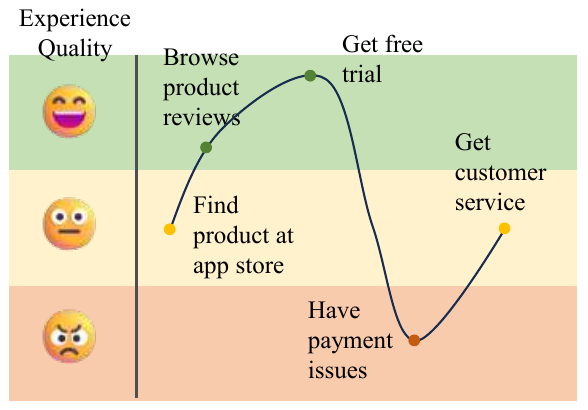}
        \caption{An example of user journey map. The content in the figure is fictional only for demonstration purpose.}
        \label{fig:user_journey_map}
    \end{minipage}
\end{figure*}

Recently, video multimodal large language models (Video MLLMs) have undergone rapid development. With increasingly powerful video understanding capabilities and support for diverse input modalities \cite{Li2024LLaVAOneVisionEV,Zhang2024InternLMXComposer25AV,Bai2025Qwen25VLTR,Chen2024Intern2_5VL,Zhang2024LongCT,Xu2025Qwen25OmniTR}, MLLMs are being deployed across a growing range of real-world scenarios.

Beyond advancements in model architecture and training paradigms, there is also a surge of interest in exploring novel interaction paradigms between users and models. A comparative illustration of these interaction methods is presented in \cref{fig:interaction}.
In offline interaction, users must upload the entire video before posing any questions, allowing the model to generate responses after consuming the whole video. In contrast, online interaction allows users to query the model in real time as the video plays, with the model required to respond immediately using only the information observed up to that point.

In addition to offline and online paradigms, proactive interaction has emerged as a promising and increasingly studied direction in video-text MLLMs \cite{Chen2024VideoLLMonlineOV,Wang2024MMDuet,Qian2025DispiderEV,Yao2025TimeChatOnline}. The defining characteristic of proactive interaction is that the model autonomously determines when to respond during video playback, rather than replies solely in response to user-initiated queries. This capability necessitates continuous monitoring of evolving visual and textual cues, real-time detection of salient moments, and timely, contextually appropriate responses.
Proactive video MLLMs hold significant potential for real-time scenarios, including live stream understanding, intelligent surveillance, egocentric assistants, and socially interactive AI agents.

While several models claim to have proactive response capabilities, their evaluations are often conducted on benchmarks that do not actually require such novel interaction. For example, most experiments are performed in offline settings where models are not required to autonomously determine when to respond, and are evaluated using multiple-choice questions rather than open-ended dialogue, which significantly differs from real-world application scenarios.

Although some benchmarks do include tasks that require response timing decisions \cite{Lin2024StreamingBenchAT,Wang2025OmniMMIAC}, they still suffer from critical limitations. Specifically: (1) the questions and answers are overly simplistic (e.g., “Inform me when [event] happens”), thereby evaluating only the timing of the response while ignoring the quality of its content; or (2) they have only a single round of response, rather than supporting multi-turn, context-aware interactions.

A more fundamental issue is the lack of appropriate evaluation metrics. Unlike in offline or online interactions, proactive interaction allows the model autonomously generating a sequence of responses at varying points in time, with the output evolving continuously. As such, evaluation methods must capture this temporal progression and take the models' timing strategy into account, rather than relying solely on static snapshots or isolated response instances.

To address this gap and advance research on proactive interaction, we introduce \benchname, the first comprehensive benchmark specifically designed to evaluate the proactive interaction capabilities of MLLMs. \benchname specifically targets the task of question answering: given a question presented at the very beginning of the video, the system must proactively detect when relevant information appears in one or more video segments and initiate responses accordingly as the video progresses. Compared to more open-ended interactive tasks, this question-answering setup offers relatively objective evaluation criteria (e.g., the availability of ground-truth answers), making it particularly suitable for academic research.
By constructing a suite of diverse tasks and collecting videos from a wide range of sources, \benchname encompasses several representative application scenarios for proactive interaction. It includes a broad spectrum of video topics, integrates multiple modalities, and supports multi-turn outputs to reflect realistic and varied use cases.

In addition, we propose \metricname (Proactive Area Under Curve), a novel evaluation metric tailored to better capture the performance of proactive interaction systems. The design of \metricname is inspired by the user journey map \cite{Huo2023AUE}, a widely used visualization tool in human-computer interaction research. An example of user journey map is shown in \cref{fig:user_journey_map}. A user journey map typically employs a line graph to illustrate how users interact with a system over time, capturing the temporal dynamics of their experiences, including shifts in emotion, engagement, and pain points—rather than relying on static snapshots.

Analogously, \metricname employs a line graph to track how the quality of a model’s responses evolves throughout the video, thereby highlighting the inherently dynamic nature of proactive interaction. This temporal perspective sets \metricname apart from traditional metrics designed for offline interactions, which evaluate only static textual outputs.

To summarize, this work presents three key contributions:

\textbf{(1) Benchmark with Multi-turn, fully open-ended answers.}
Unlike most existing video question answering benchmarks that predominantly adopt multiple-choice formats, \benchname features fully open-ended questions requiring multi-turn, free-form textual responses. While this design introduces additional challenges for evaluation, it offers a more realistic and comprehensive assessment of a model’s interactive capabilities and is closer to real-world application scenarios.

\textbf{(2) Benchmark with diverse topics and multimodal inputs.}
\benchname includes videos covering a wide array of topics that are highly relevant to proactive interaction, such as web videos, egocentric recordings, television series, and surveillance footage. It also incorporates multiple input modalities including text, video, and speech, reflecting the complexity and richness of real-world use cases.

\textbf{(3) A reply time-aware metric for proactive evaluation.}
We introduce \metricname, a novel evaluation metric that captures the evolving nature of model responses in proactive interaction settings. By explicitly modeling how response quality changes over time, \metricname provides stronger alignment with human judgments and better reflects the user experience.
\section{Related Works}

\subsection{Video Understanding Benchmarks}
Recent years have witnessed a surge in video understanding and question-answering benchmarks. \cite{Li2023MVBenchAC,Fu2024VideoMMETF,Cai2024TemporalBenchBF,Li2023VITATECSAD,Li2023SEEDBenchBM,Li2024SEEDBench2PlusBM,Fang2024MMBenchVideoAL}
While these benchmarks cover videos of diverse topics, lengths, and question-answering formats, the interaction method remains predominantly in the most widely studied offline interaction.

To explore the applicability of video-text MLLMs in streaming video scenarios, a number of benchmarks have recently been introduced, often labeled with terms such as ``\textit{streaming}'' or ``\textit{online}'' \cite{Lin2024StreamingBenchAT,Wang2025OmniMMIAC,Li2025OVOBenchHF,liu2024etbenchopenendedeventlevel}. These terms emphasize that user queries are injected at specific time points during video playback. However, they generally do not enable any flexibility on when the model should respond. In fact, with the exception of a few sub-tasks, the majority of these benchmarks require the model to respond immediately following a user question, effectively reduces the task to offline video understanding up to the time of the query.

Distinct from all prior works, this study introduces the first benchmark and evaluation metric specifically designed for proactive interaction, explicitly accounting for the temporal evolution of model responses. Moreover, it emphasizes open-ended question answering, despite posing greater challenges for evaluation, better reflects the demands of real-world applications compared to multiple-choice formats.

\subsection{Proactive Video-Text LLMs}
VideoLLM-Online \cite{Chen2024VideoLLMonlineOV} is one of the first works that adapt video-text MLLMs to proactive interaction scenarios. MMDuet \cite{Wang2024MMDuet} improves upon this by being trained on more diverse tasks and datasets, yet it still face problems like inaccurate response timing and redundant outputs. Dispider \cite{Qian2025DispiderEV} introduces a disentangled framework of perception, decision, and reaction, and TimeChat-Online \cite{Yao2025TimeChatOnline} focuses on the token compression techniques of the input video stream.

While these studies propose different approaches to enhance proactive modeling, the majority of their experiments are conducted under non-proactive interaction settings, where models are not required to autonomously determine response timing. Consequently, they fall short of thoroughly evaluating the core capabilities needed for proactive interaction. This underscores the urgent need for a benchmark specifically designed to evaluate and facilitate the development of proactive video-text MLLMs.

\section{The \metricname Metric} \label{sec:metric}

In existing NLP evaluation methods, metrics are typically computed by comparing the model’s complete output against the ground-truth answer, using criteria such as $n$-gram overlap \cite{Papineni2002BleuAM,Vedantam2014CIDErCI}, semantic similarity \cite{Zhang2019BERTScoreET} or LLM-based Evaluation \cite{Li2024LLMsasJudgesAC}. While these metrics are widely used to assess the quality of textual outputs, they fall short in capturing how model performance evolves over time in proactive interaction scenarios.

To address this limitation, we propose \metricname (Proactive Area Under Curve), a novel evaluation metric that jointly considers both the timing and content of model responses within a unified framework. Drawing inspiration from the concept of a user journey map which visualizes user experience as a dynamic line graph over the course of interactions, \metricname{} plots a timestamp-score curve based on the model’s outputs and computes the area under the resulting polyline to represent the model’s proactive capabilities.

This design enables \metricname{} to reflect the temporal evolution of user experience (i.e., the correctness of model responses over time), which is a defining feature of proactive interaction compared to traditional offline interaction.

Formally, suppose there are \textbf{$G$ turns of ground-truth replies} in a video, where each reply consists of a textual content $gold_g$ and an associated timespan $(t_g^{start}, t_g^{end})$, for $g = 1, 2, \dots, G$. This indicates that during the interval $(t_g^{start}, t_g^{end})$, the user expects to receive the information contained in $gold_g$. While we acknowledge that evaluating proactive interaction experiences is inherently subjective, here we assume the existence of an ideal ground truth for a quantitative and objective measurement of proactive model performance.

\textbf{\metricname{} operates independently on each ground-truth reply turn.} Since the following introduction to \metricname is conducted within a single reply turn $(gold_g, t_g^{start}, t_g^{end})$, we will omit the subscript ``$g$'' for simplicity.

Suppose that, for a given reply timespan $(t^{start}, t^{end})$, there are $P$ model responses that fall within this interval. Each response is associated with textual content $pred_{p}$ and a timestamp $\tau_{p}$, where $p = 1, 2, \dots, P$ and $t^{start} < \tau_{1} < \tau_{2} < \dots < \tau_{P} < t^{end}$.

To assess the correctness of model predictions up to each timestamp $\tau_{p}$, we input the question, the ground-truth answer $gold$, and the set of model responses generated before $\tau_{p}$, i.e., $\{pred_{1}, pred_{2}, \dots, pred_{p}\}$, into a large language model (GPT-4.1 in our implementation). The model is instructed to assign a score reflecting how well this set of accumulated responses aligns with the ground-truth answer. This score is denoted as $s_{p}$, representing the quality of the model’s responses up to $\tau_{p}$.

In our implementation, $s_{p}$ takes a discrete value from ${0, 1, 2}$ (with a maximum score $S = 2$), corresponding to completely incorrect, partially correct, and mostly correct predictions, respectively. We also experimented with finer-grained scoring scales (e.g., ${0, 1, 2, 3, 4}$ with $S = 4$). However, human studies and interviews revealed that evaluators are generally insensitive to subtle differences in response quality under proactive interaction settings. Consequently, we adopt a coarser-grained scale with $S = 2$ for better consistency and interpretability.

Finally, we construct a polyline in the time-score coordinate space by using $\tau_{p}$ as the $x$-axis values and the corresponding $s_{p}$ as the $y$-axis values. To make the polyline continuous in $(t^{start}, t^{end})$, we add two additional points as endpoints of the polyline: $(t^{start}, 0.5)$ as the initial point and $(t^{end}, s_{P})$ as the final point. The initial score of 0.5 reflects the intuition that providing no response is preferable to giving entirely incorrect answers, which receive a score of 0 from the LLM evaluator. The final \metricname score for this ground-truth reply turn is defined as the ratio of the area under this curve to the maximum possible area ( by $(t^{end} - t^{start}) \times S$) as calculated by \cref{eq:auc}, where $S$ is the maximum score.

\begin{equation}\label{eq:auc}
\begin{aligned}
    & \metricname = [ (\tau_{1}-t^{start}) \times 0.5 + \\
    & \sum^{P-1}_{p=1}{(\tau_{p+1}-\tau_{p}) \times s_{p}} + (t^{end}-\tau_{P}) \times s_{P} ] \\
    & \div (q^{end}-q^{start}) \times S
\end{aligned}
\end{equation}

We use \cref{eq:auc} to calculate $\metricname_g$ for each ground truth reply turn $(gold_g, t_g^{start}, t_g^{end})$, and use the average value of all turns as the final \metricname score of the entire video.

\begin{figure}
    \centering
    \includegraphics[width=\linewidth]{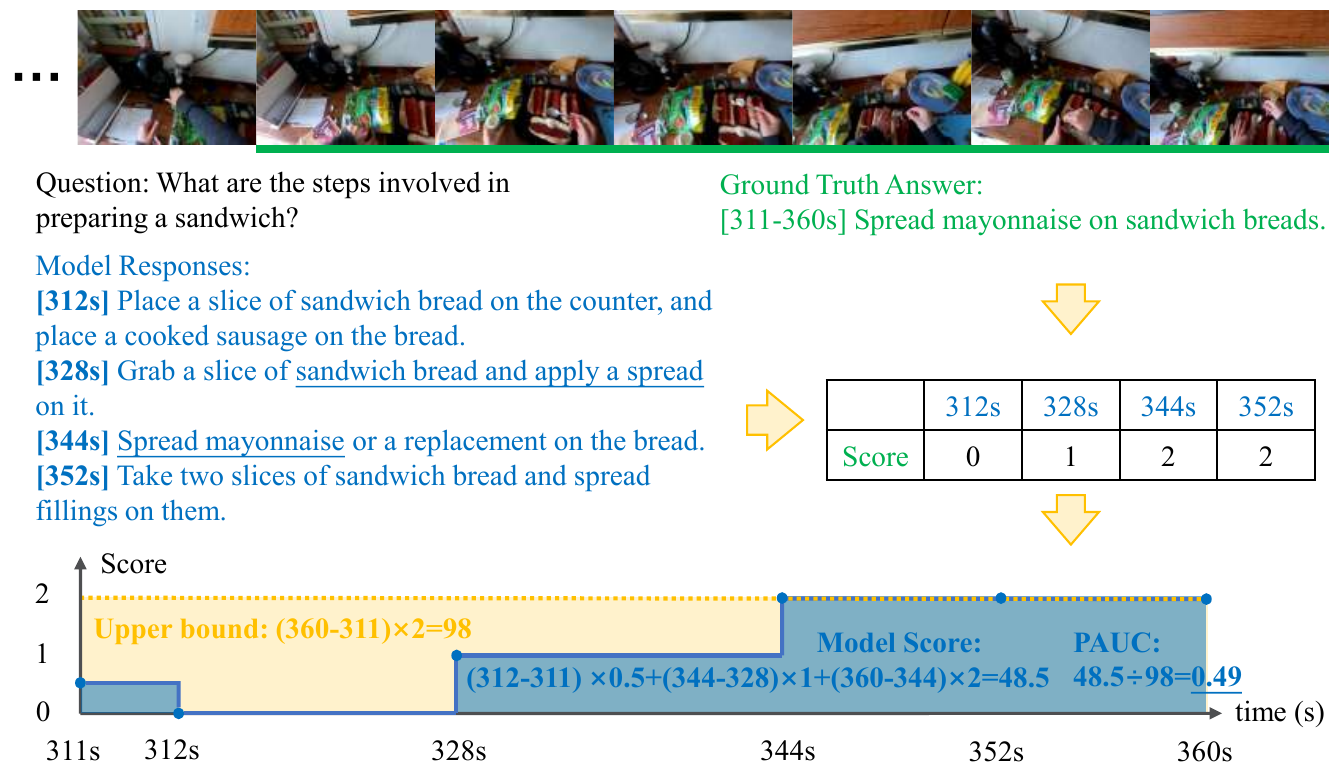}
    \caption{An Illustration of the \metricname metric.}
    \label{fig:metric}
\end{figure}

Despite its simplicity and intuitive design, the computation method of \metricname{} effectively achieves several key goals:

First, it rewards model responses that are both early and accurate. The more closely a model's reply aligns with the ground-truth answer, the higher the score assigned by the LLM evaluator, thereby increasing the overall \metricname{} value. Furthermore, when the correctness of a reply remains constant (e.g., score $= 2$), earlier delivery leads to an earlier rise in the score curve. This results in a larger area under the curve and consequently a higher \metricname{} score.

Second, \metricname{} penalizes incorrect responses. If the model produces a reply that contradicts the ground-truth answer, this incorrect response is included in the accumulated input to the LLM for all subsequent timestamps. As a result, the presence of an incorrect reply reduces the likelihood of receiving high scores at later points, thereby lowering the area under the curve and the final \metricname{} value.

\subsection{Adjusting the Importance of Timeliness}
However, in real-world applications, different tasks often impose different demands on the importance of timeliness. For example, in scenarios where timeliness is critical, responses generated just before $t^{end}$ contribute only marginal value. In contrast, in tasks where textual content takes precedence over response time, even late but accurate replies arriving just before $t^{end}$ can still be highly valuable.

To accommodate these differing priorities, we introduce a hyperparameter $\omega \in [0, 1]$ to balance the importance of timeliness and correctness. Intuitively, $\omega$ controls the extent to which the $x$-coordinates ($\tau_{p}$) of the points on the polyline are shifted leftward along the time axis: $\tau_{p} \to \tau_{p}^\prime = t^{start} + (1 - \omega) \times (\tau_{p} - t^{start})$.

When $\omega = 0$, we have $\tau_{p}^\prime = \tau_{p}$, meaning that the $x$-coordinates are not shifted leftward. This setting reflects scenarios where timeliness is very important: when a response’s timestamp $\tau_{p}$ approaches the end of the reply timespan of the turn $t^{end}$ (i.e., $t^{end} - \tau_{p} \to 0$), its contribution to the final score becomes negligible according to \cref{eq:auc}.

In contrast, when $\omega > 0$, all $x$-coordinates are shifted left proportionally to their distance from $t^{start}$, thereby compressing the time intervals between adjacent predictions $(\tau_{p} - \tau_{p-1})$. This reduces the influence of temporal differences among responses. Meanwhile, as the gap between the last response time and end time, namely $t^{end} - \tau_{P}$, increases, the correctness score computed from all accumulated responses (recall that $s_{P}$ is obtained using the entire set of replies within the interval, $\{pred_{1}, pred_{2}, \dots, pred_{P}\}$) exerts a greater effect on the final \metricname value. This behavior corresponds to scenarios where correctness increasingly outweighs timeliness as $\omega$ grows. In the extreme case of $\omega = 1$, all keypoints’ $x$-coordinates are shifted entirely to $t^{start}$. Here, \cref{eq:auc} degenerates to $(t^{end} - t^{start}) \times s_{P}$, equivalent to directly evaluating the correctness of the concatenated responses while completely ignoring their reply times. Here we recommand using $\omega=0.5$ as the default setting.

It is also worth emphasizing that the \metricname{} framework is highly flexible. The use of an LLM as the evaluator (i.e., generating the correctness score on the $y$-axis) can be replaced with alternative metrics such as BLEU \cite{Papineni2002BleuAM}, CIDEr \cite{Vedantam2014CIDErCI}, or accuracy, depending on specific application requirements.

\begin{figure}
    \centering
    \includegraphics[width=\linewidth]{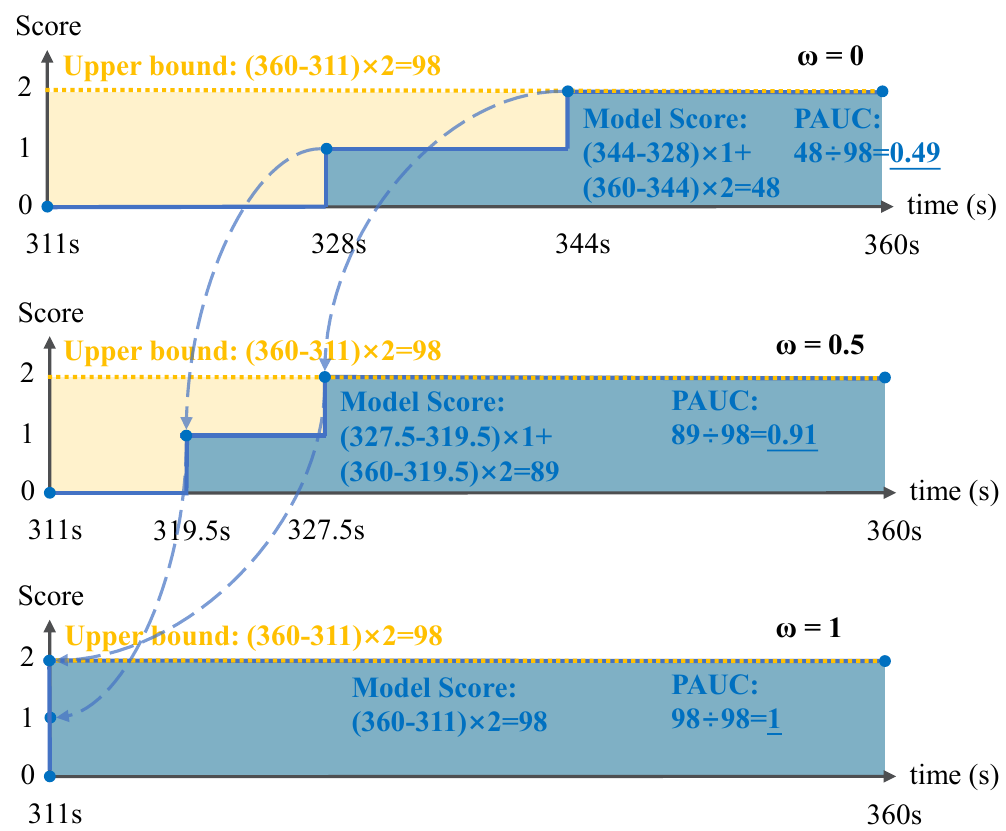}
    \caption{An Illustration of the effects of $\omega$.}
    \label{fig:omega}
\end{figure}

\section{The \benchname Benchmark}
\begin{figure*}
    \centering
    \includegraphics[width=\linewidth]{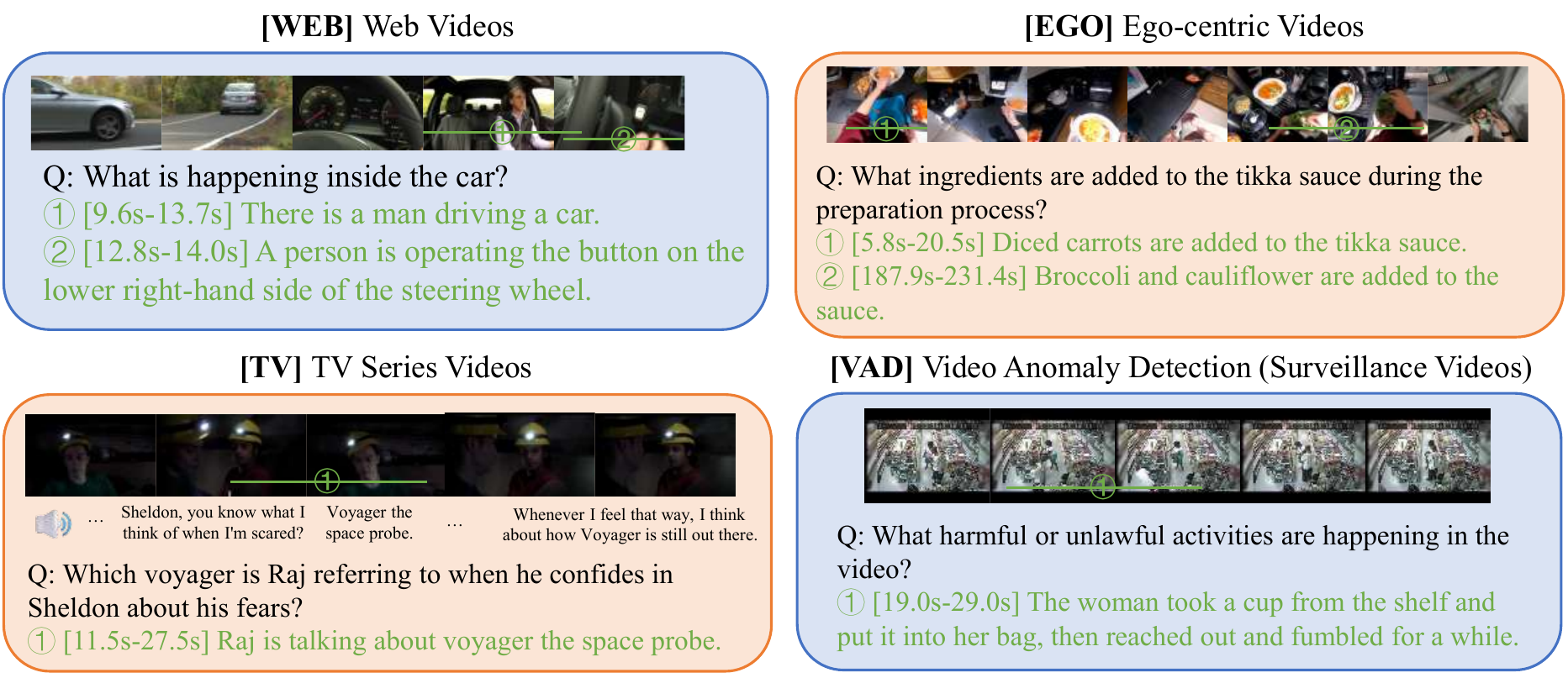}
    \caption{Example data from different tasks.}
    \label{fig:examples}
\end{figure*}

To make use of \metricname, a benchmark is required for evaluating proactive models. We introduce \benchname, the first comprehensive benchmark designed for proactive interaction. To encompass prevalent scenarios in proactive interaction, \benchname focuses on four key tasks:

(1) \textbf{proactive web-video QA} (\texttt{[WEB]}): centering on general web-video understanding.
(2) \textbf{proactive ego-centric video QA} (\texttt{[EGO]}): centering on first-person-view video comprehension, particularly relevant in robotics and daily assistant applications.
(3) \textbf{proactive TV-series video QA} (\texttt{[TV]}): emphasizing dialogue and social relationship understanding with speech input, and
(4) \textbf{proactive video anomaly detection} (\texttt{[VAD]}) targeting surveillance video monitoring and alerting.
Illustrative examples of these tasks are provided in \cref{fig:examples}, while a detailed dataset statistics is presented in \cref{tab:stat_tasks,tab:stat}.

\subsection{Dataset Construction}

\subsubsection{Data Source}
There are already many offline VideoQA benchmarks focusing on various capabilities. To construct \benchname, we source video and annotations from Shot2story-MAGQA-39k for \texttt{[WEB]} \cite{Wang2024MMDuet,Han2023Shot2StoryAN}, Ego4D Goalstep \cite{Song2023Ego4DGT} for \texttt{[EGO]}, TVQA \cite{Lei2018TVQALC} for \texttt{[TV]}, and UCF-Crime \cite{Sultani2018RealWorldAD,Yuan2023TowardsSV} for \texttt{[VAD]}.

\subsubsection{Question and Answers in \benchname}
For Shot2story-MAGQA-39k and TVQA, questions, answers, and relevant timespans are already provided, we directly use these annotations as model input. For Ego4D Goalstep only dense video descriptions are provided, to create questions and answers we follow the pipeline of \cite{Wang2024MMDuet} to generate QAs from dense captions. For the \texttt{[VAD]} task, as UCF-Crime only contains timespans but not the textual descriptions of the anomaly event, we manually write a description for each anomaly event as the answer and use ``What suspicious or harmful activities, including unlawful, criminal behaviors or destructive accidents, are happening in the video?'' as the question. 
For all datasets, if two consecutive ground truth turns have similar textual contents (judged by LLM and text overlap) and the interval between their reply timespans is less than 3 seconds, we merge them into one ground truth turn, as the two scenes are likely describing the same action.

\begin{table}[]
    \centering
    \setlength{\tabcolsep}{3pt}
    \begin{tabular}{c|c|c|c|c}
        \toprule
         & \texttt{[WEB]} & \texttt{[EGO]} & \texttt{[TV]} & \texttt{[VAD]} \\
        \hline
        \# videos & 500 & 326 & 450 & 101 \\
        \# examples & 500 & 326 & 500 & 101 \\
        \# reply turns & 1328 & 1575 & 500 & 107 \\
        reply / example & 2.66 & 4.83 & 1.00 & 1.06 \\
        video len (s) & 16.59 & 360.00 & 75.57 & 121.03 \\
        reply span len (s) & 5.51 & 29.20 & 12.08 & 17.96 \\
        \bottomrule
    \end{tabular}
    \caption{Dataset Statistics for each task from \benchname.}
    \label{tab:stat_tasks}
\end{table}

\begin{table*}[]
    \normalsize
    \centering
    \setlength{\tabcolsep}{3pt}
    \begin{tabular}{c||c|c|c|c|c|c}
        \toprule
        Benchmark & Modalities & \#Videos  & \#Questions & \makecell{Multi- \\ Answer} & \makecell{Open- \\ Ended} & Proactive \\
        \hline
        MVBench \cite{Li2023MVBenchAC} & Video & 3,641 & 4,000 & \usym{2717} & \usym{2717} & \usym{2717} \\
        VideoMME \cite{Fu2024VideoMMETF} & Video, Audio & 900 & 2,700 & \usym{2717} & \usym{2717} & \usym{2717} \\
        OmniBench \cite{Li2024OmniBenchTT} & Image, Audio & - & 1,142 & \usym{2717} & \usym{2717} & \usym{2717} \\
        OVO-Bench \cite{Li2025OVOBenchHF} & Video & 644 & 2,814 & \usym{2717} & \usym{2717} & \usym{2713}$^\ast$ \\
        StreamingBench \cite{Lin2024StreamingBenchAT} & Video, Audio &  900 & 4,500 & \usym{2717} & \usym{2717} & \usym{2713}$^\ast$ \\
        OmnniMMI \cite{Wang2025OmniMMIAC} & Video, Audio &  1,121 & 2,290 & \usym{2713}$^\ast$ & \usym{2713}$^\ast$ & \usym{2713}$^\ast$ \\
        \hline
        \benchname (Ours) & Video, Audio & 1,377 & 1,427 & \usym{2713}$^\ast$ & \usym{2713} & \usym{2713} \\
        \bottomrule
    \end{tabular}
    \caption{Comparison with existing Video Benchmarks. \protect\usym{2713}$^\ast$: True for some of the sub-tasks in the benchmark}
    \label{tab:stat}
\end{table*}

\section{Employing Offline Video-Text LLMs for Proactive Interaction}

To the best of our knowledge, as of the writing of this manuscript, only two video-text LLMs designed for proactive interaction, VideoLLM-Online~\cite{Chen2024VideoLLMonlineOV} and MMDuet~\cite{Wang2024MMDuet}, have fully open-sourced their code to support proactive evaluation.
To enable the evaluation of a broader set of models on \benchname, we employ a simple rule-based strategy to adapt offline video-text LLMs for proactive interaction. Specifically, we segment each video into fixed-length chunks and, at each timestep, provide the model with the current video chunk, the associated question, and the model’s previous response as input. The model is first required to determine whether the current video chunk can answer the question, and output one of the following: “I have no answer,” “I have the same answer” (as the previous response), or “I have a new answer.” If the model responds with “I have a new answer,” it is then required to generate the updated response accordingly.
In practice, we find that only proprietary models are capable of reliably following these multi-step instructions. Open-source models, by contrast, typically fail to comply and result in unexpected behaviors (e.g., ignore the instructions and start answering questions directly). Consequently, for open-source models, we adopt a simplified strategy: each video chunk is presented to the model alongside the question, and the model is asked to determine whether the chunk contains sufficient information to answer the question. If the model responds affirmatively, we perform an additional round of inference to obtain the answer based on the current chunk.

We also explored alternative strategies for adapting offline models to proactive interaction, such as incrementally increasing the number of video chunks provided as input. However, these approaches did not yield satisfactory results. This highlights the significant challenges involved in applying existing offline video-text MLLMs to proactive interaction without targeted training. Further exploration in this direction, such as leveraging agent-based systems, remains an important direction for future research. In this paper, we present only a subset of possible solutions based on our current understanding. Additional experimental details can be found in the Appendix.

\section{Experiments}

\begin{table*}[]
    \centering
    \setlength{\tabcolsep}{3pt}
    \begin{tabular}{c|c|c|c|c|c|c|c|c|c|c|c|c}
        \toprule
        Model & \multicolumn{3}{|c|}{\texttt{[WEB]}} & \multicolumn{3}{c|}{\texttt{[EGO]}} & \multicolumn{3}{c|}{\texttt{[TV]}} & \multicolumn{3}{c}{\texttt{[VAD]}} \\
        \hline
        $\omega=$ & 0.0 & 0.5 & 1.0 & 0.0 & 0.5 & 1.0 & 0.0 & 0.5 & 1.0 & 0.0 & 0.5 & 1.0 \\ 
        \hline
        \hline
        Human & 36.5 & 38.6 & 40.7 & 35.0 & 38.2 & 41.3 & 38.1 & 47.0 & 55.9 & 47.4 & 53.6 & 59.8 \\ 
        
        \hline
        \multicolumn{3}{l}{\textit{Proprietary Offline Models}} \\
        \hline
        GPT-4.1 & 44.6 & 51.7 & 58.9 & 53.6 & 58.8 & 64.0 & 45.0 & 56.8 & 68.5 & 40.8 & 46.2 & 51.6 \\
        GPT-4.1-mini & 41.2 & 47.8 & 54.5 & 59.1 & 65.8 & 72.5 & 48.5 & 59.4 & 70.3 & 41.4 & 47.7 & 54.0 \\
        Gemini-1.5-pro & 37.1 & 42.1 & 47.0 & 47.0 & 49.7 & 52.4 & 41.7 & 52.0 & 62.4 & 34.3 & 36.1 & 37.9 \\
        Gemini-2.0-flash & 36.1 & 41.0 & 45.9 & 49.4 & 53.7 & 57.9 & 40.4 & 49.1 & 57.8 & 32.9 & 35.8 & 38.6 \\

        \hline
        \multicolumn{3}{l}{\textit{Open-Sourced Offline Models}} \\
        \hline
        InternVL-2.5 8B & 41.6 & 48.2 & 54.9 & 52.1 & 57.8 & 63.6 & 36.5 & 41.5 & 46.6 & 22.2 & 21.5 & 20.8 \\
        LLaVA-OV 7B & 46.6 & 55.0 & 63.4 & 57.0 & 61.6 & 66.1 & 38.2 & 45.1 & 51.9 & 25.3 & 25.6 & 25.9  \\ 
        LongVA 7B & 39.3 & 47.2 & 55.1 & 34.5 & 37.4 & 40.2 & 35.3 & 41.5 & 47.6 & 27.2 & 29.8 & 32.3 \\
        Qwen2.5-VL 7B & 45.7 & 52.7 & 59.8 & 42.8 & 46.5 & 50.3 & 32.7 & 36.5 & 40.2 & 27.9 & 29.3 & 30.7 \\
        IXC-2.5 7B & 43.0 & 50.3 & 57.6 & 43.0 & 50.3 & 57.6 & 41.0 & 48.5 & 56.0 & 25.4 & 26.8 & 28.3 \\
        \hline
        \multicolumn{3}{l}{\textit{Proactive Models}} \\
        \hline
        MMDuet & 37.2 & 38.9 & 40.7 & 44.0 & 46.0 & 47.9 & 20.7 & 21.1 & 21.6 & 26.4 & 27.4 & 28.5 \\
        MMDuet+rm.ass.turns & 41.4 & 43.5 & 45.6 & 49.4 & 52.2 & 55.0 & 28.2 & 32.6 & 37.1 & 38.5 & 42.5 & 46.5 \\
        VideoLLM-Online & 25.9 & 25.9 & 25.9 & 25.0 & 25.0 & 25.1 & 17.8 & 18.3 & 18.8 & 25.0 & 25.0 & 25.0 \\

        \bottomrule
    \end{tabular}
    \caption{Results on \benchname with different $\omega$.}
    \label{tab:main_results}
\end{table*}

\begin{table}[]
    \centering
    \begin{tabular}{c|c|c||c}
    \toprule
    \multirow{2}{*}{Task} & \multicolumn{3}{c}{Agreement w/ Human} \\
     & $\omega=1$ & $\omega=0.5$ & Human \\
    \hline
    \texttt{[WEB]} & 0.23/0.30 & 0.37/0.40 & 0.50/0.49 \\ 
    \texttt{[EGO]} & 0.26/0.32 & 0.30/0.35 & 0.34/0.31 \\ 
    \texttt{[TV]} & 0.29/0.37 & 0.34/0.37 & 0.55/0.59 \\ 
    \texttt{[VAD]} & 0.31/0.36 & 0.45/0.49 & 0.40/0.51 \\
    \bottomrule
    \end{tabular}
    \caption{Agreement between human preference and \metricname $\omega=1$ (baseline metric without taking reply time into account), and \metricname $\omega=0.5$ (taking reply time into account). We also report the agreement between 2 different sets of annotators as a reference. The agreement metrics reported are Cohen's kappa with no-weighting/linear-weighting \cite{Cohen1960ACO}}
    \label{tab:human_alignment}
\end{table}

\begin{table}[]
    \centering
    \setlength{\tabcolsep}{3pt}
    \begin{tabular}{c|c|c|c|c}
        \toprule
        Model & \texttt{[WEB]} & \texttt{[EGO]} & \texttt{[TV]} & \texttt{[VAD]} \\
        \hline
        MMDuet & 81.3 & 99.4 & 92.8 & 99.2 \\
        \hline
        \makecell{MMDuet w/ \\ rm. ass. turns} & 81.1 & 92.6 & 61.2 & 80.9 \\
        \hline
        \makecell{VideoLLM- \\ Online$^\dag$} & - & - & 53.9 & - \\
        \bottomrule
    \end{tabular}
    \caption{The proportion of duplicate pred turns to all pred turns (excluding the first pred turn in each ground-truth answer turn). $\dag$: Videollm-online generated more than 1 reply for only less than 10 answer turns on the \texttt{[WEB]}, \texttt{[EGO]}, and \texttt{[VAD]} datasets. Since the sample size is too small, we are not reporting this result as they have overly-large variance.}
    \label{tab:repetition}
\end{table}

We report \metricname metric on \benchname for the following methods: (1) proprietary offline video MLLMs, (2) open-sourced offline video MLLMs, (3) open-sourced proactive video MLLMs, and (4) human performance.

For offline models, we use a video chunk size of 2 seconds for \texttt{[WEB]} and 5 seconds for other datasets. We sample 2 frames per second for \texttt{[WEB]} and 1 frame per second for other datasets.
For the baseline models that do not accept audio input, for the \texttt{[TV]} task we input the text-form subtitles to the model at the beginning timestamp of an utterance in the TV Series.
For human performance, we recruit 4 human annotators, instruct them to read the question before watching the video, pause the video every time when it plays to the segment where the question can be answered, and write down the current video timestamp along with an answer.
Since completing this task manually is very labor-intensive, we only sampled 60 videos from each dataset to evaluate human performance.

\subsection{Main Results}
The results are listed in \cref{tab:main_results}. We have the following observations:

(1) Human performance is relatively low, largely due to the demanding nature of the task. Annotators are required to pause the video and write responses precisely when answer-relevant segments appear, which is both cumbersome and unnatural. In practice, many annotators tend to provide responses retrospectively, sometimes even after the relevant reply timespan has passed, rather than offering a brief, timely response followed by iterative refinements, as the model is designed to do, despite explicit instructions to the contrary.
Moreover, certain dataset characteristics further increase the difficulty for human annotators. For example, the reply timespans in \texttt{[WEB]} are typically short, while the \texttt{[EGO]} dataset contains a large number of ground-truth turns per video, making it challenging to label answers accurately and in detail.
Finally, since the annotators are not native English speakers, the annotation process was conducted in Chinese with back-and-forth translation. This additional step may have introduced translation-related ambiguities or delays, contributing to further degradation in human performance.

(2) On \texttt{[TV]} and \texttt{[VAD]} tasks, proprietary models significantly outperform both open-source and proactive models. This performance gap can be attributed to the complexity of these tasks, which require deep understanding of video content, such as character relationships and subtitles in \texttt{[TV]}, or perceptually demanding surveillance footage in \texttt{[VAD]}.
In contrast, the advantage of proprietary models is less evident on the \texttt{[WEB]} and \texttt{[EGO]} tasks. In these benchmarks, the core challenge lies in accurately determining the timing of responses, a capability that remains underdeveloped across all models, including those explicitly designed for proactive interaction.

(3) Proactive models do not demonstrate better results than offline models. This is because although these models can theoretically make decisions on response timing, due by relatively simple training techniques and limited training resources, their performance is not satisfactory. Our further analysis in \cref{tab:repetition} confirms that these models tend to repeat previously generated content, leading to lower response quality.

\subsection{Alignment with Human Preferences}
To validate the effectiveness of the proposed \metricname metric, we conduct a human study to assess its agreement with human preferences at ground-truth reply turn level. Specifically, we sample 100 ground-truth reply turns from each task (and 50 answer turns from \texttt{[VAD]}, due to its smaller dataset size), and collect two model predictions per sample using the \textit{Incremental Chunks} method from GPT-4.1-mini and Gemini-2.0-Flash. Human annotators are then asked to indicate their preference between the two predictions (one model prediction wins or draws).
To ensure the informativeness of the evaluated responses, we apply the following sampling criteria: (1) both models must produce at least one response within the reply span, and at least one of them must respond in more than one round; and (2) both models must have at least one response with a \metricname score greater than 0.

Annotators are instructed to assume the role of users seeking timely and accurate information from the video. They are asked to judge which model prediction better captures the information present in the ground truth at earlier timestamps, while also considering text quality (e.g., avoiding hallucinations and maintaining fluency). To assess inter-annotator consistency, 50 examples for each task are annotated twice independently by two different annotators. 
The results are presented in \cref{tab:human_alignment}. Compared to the baseline metric that does not account for reply timing ($\omega=1$), the proposed \metricname metric which incorporates reply time with $\omega=0.5$ consistently exhibits stronger alignment with human preferences and approaches the level of agreement observed between human annotators.

Nonetheless, all Cohen’s kappa scores, including the agreement between human annotators, are relatively low. Several factors contribute to this outcome: (1) the trade-off between correctness and timeliness is inherently subjective, and it is natural for different individuals to prioritize these aspects differently; and (2) the examples selected for human preference annotation represent the more complex cases. Compared to easier instances where one model's response clearly outperforms the other's, focusing on these borderline cases provides a more effective test of the proposed \metricname metric.

To intuitively demonstrate the advantages of \metricname compared to the baseline metric on the addition of temporal changes, we present several examples as case study in the appendix.

\section{Conclusion}
In this work, we address the emerging challenge of evaluating proactive multimodal dialogue systems by introducing \benchname, the first comprehensive benchmark for assessing systems' capabilities in proactive interaction scenarios. We propose \metricname, a novel evaluation metric that tracks how the quality of a model’s responses evolves throughout the video, as a more valid metric to evaluate models for proactive interaction. We believe that \benchname and \metricname will serve as valuable tools for future research and development in proactive interaction models that can have potentially high-impact application areas such as live stream understanding, video anomaly detection, and ego-centric agents.



\bibliography{acl_latex}

\appendix
\section{More approches Evaluating Offline Models}
\subsection{Gradually Increasing Number of Chunks}

We also experimented with a very intuitive and efficient method for using offline models for proactive interaction: during the $n^{th}$ round of interaction, we input the first $n$ video chunks along with all the model’s prior responses from the previous $n-1$ rounds but remove the final EOS token, and observe whether the model would continue generating. If the model generates some new text (considering that existing outputs were generated conditioning on the previous $n-1$ video chunks), we could attribute this new content to the inclusion of the $n^{th}$ chunk and thus treat it as the model’s response at the end time of the $n^{th}$ video chunk.  
However, in our experiments we found that in almost all cases existing open-source models only generate answers in the first interaction round for each video. In subsequent rounds the models almost never extended their output and simply emitted an EOS token to end their turn instead.
We speculate that the underlying reason is in proactive interaction, whether the model should initiate a response is largely determined by the newly added video segment, and current open-source models were not trained for this.

\section{Case Study}
To intuitively demonstrate the advantages of \metricname compared to the baseline metric that does not account for temporal changes, here we present 2 real examples tested on GPT-4.1-mini and Gemini-2.0-flash.
Since the scores are counted independently in each ground truth span, and also to simplify the information for easier display, Each example shows a case from one ground truth reply span within an example.

In the first example, Gemini-2.0-Flash responded faster and covered more key information in the ground truth answer, thus reaching the maximum score of 2 as early as 10 seconds. This gives it an advantage over GPT-4.1-mini in the final \metricname evaluation.

The second example is a more representative case where \metricname is better than time-agnostic baseline metrics for evaluating proactive systems. Both models eventually answered "stealing items", which would yield equal scores under a time-agnostic metric. However, Gemini-2.0-Flash correctly provided this information earlier at 20s and gained 1 point, while GPT-4.1-mini gave an incorrect response at the same timestamp and received 0 points. This results in a significantly higher \metricname score for the Gemini-2.0-Flash.

\begin{figure}
    \centering
    \includegraphics[width=\linewidth]{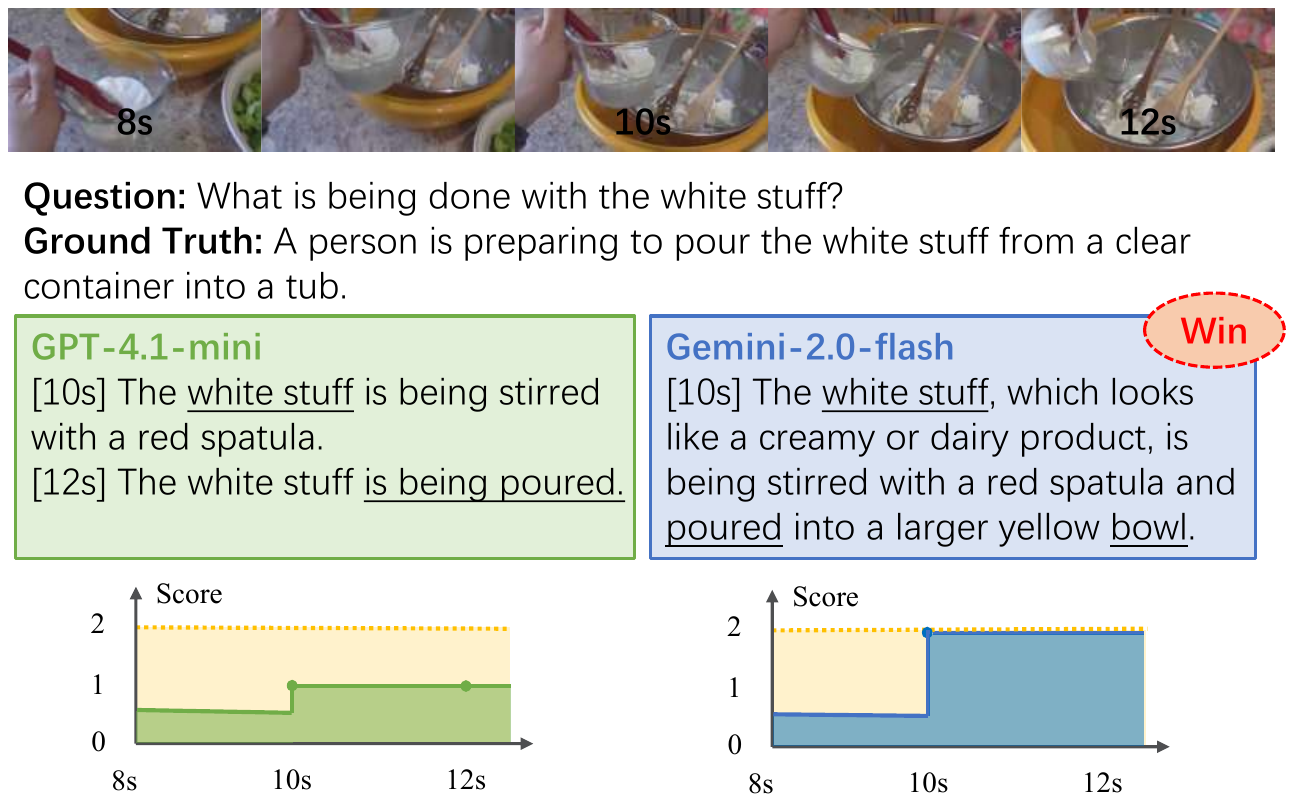}
    \includegraphics[width=\linewidth]{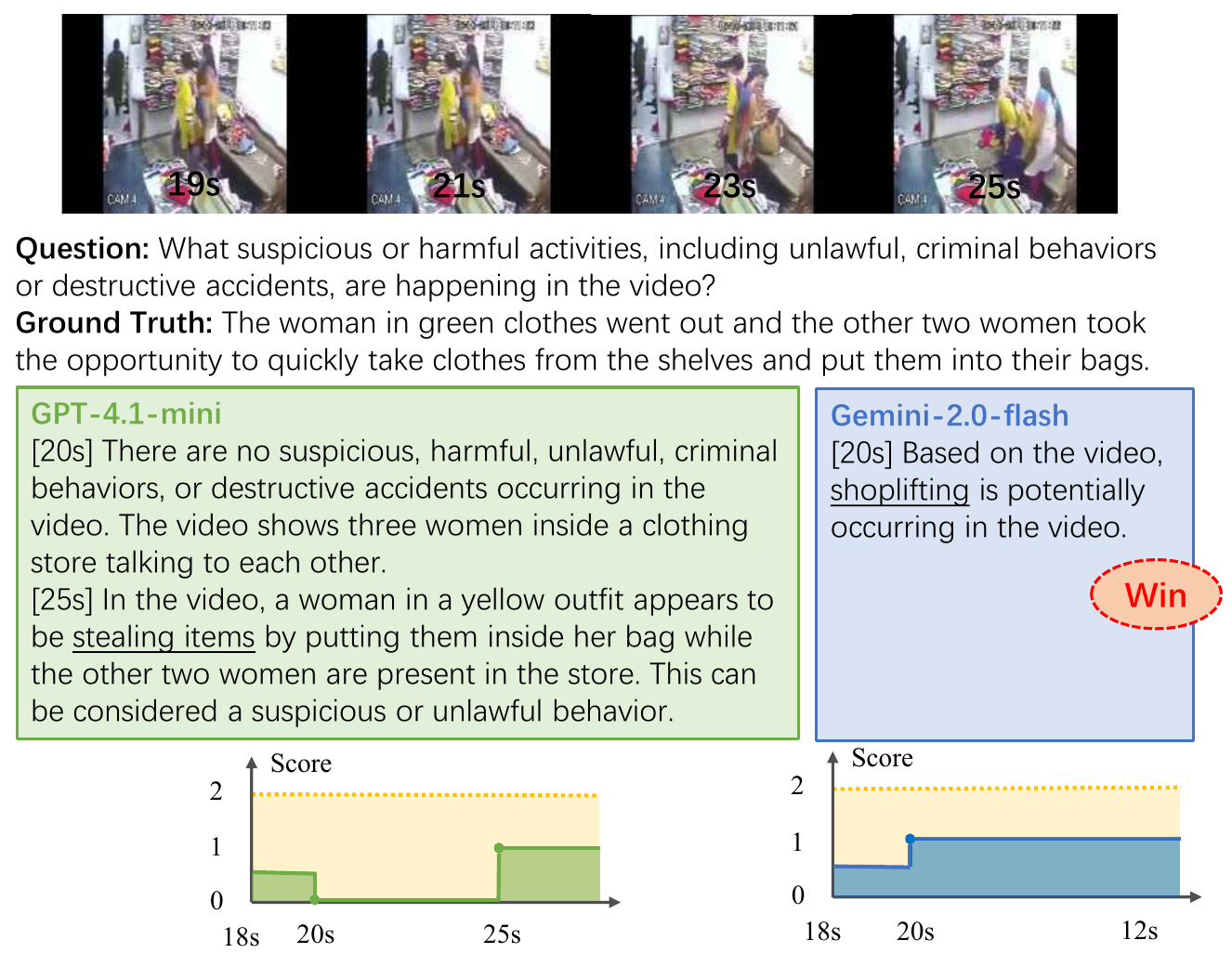}
    \caption{Qualitative demonstration of \benchname and \metricname. The first example is from [WEB], and the second example is from [VAD]. The underlined content in the model responses highlights key information (for visualization purposes only, not as part of the evaluation criteria). In the line chart below, each dot represents a model response and its impact on the score poly-line.}
    \label{fig:enter-label}
\end{figure}

\end{document}